\def\year{2021}\relax
\documentclass[letterpaper]{article} 
\usepackage{aaai21}  
\usepackage{times}  
\usepackage{helvet} 
\usepackage{courier}  
\usepackage[hyphens]{url}  
\usepackage{graphicx} 
\def\UrlFont{\rm}  
\usepackage{natbib}  
\usepackage{caption} 
\usepackage[utf8]{inputenc} 
\usepackage{booktabs}       
\usepackage{amsfonts}       
\usepackage{nicefrac}       
\usepackage{microtype}      
\usepackage{graphicx}
\usepackage{color}
\usepackage{subfig}
\usepackage{comment}
\usepackage{mathtools}
\usepackage[switch]{lineno}

\newcommand{\myparagraph}[1]{\vspace{6pt}\noindent\textbf{#1.}}

\title{Augmenting Policy Learning with \\ Routines Discovered from a Single Demonstration}

\author{
    Zelin Zhao \footnote{Work was done when Zelin was a visiting student at MIT.} \textsuperscript{\rm 1} , Chuang Gan \textsuperscript{\rm 2}, Jiajun Wu \textsuperscript{\rm 3}, Xiaoxiao Guo \textsuperscript{\rm 2}, Joshua Tenenbaum \textsuperscript{\rm 4} \\
}

\affiliations{
    \textsuperscript{\rm 1} Shanghai Jiao Tong University \\ \textsuperscript{\rm 2} MIT-IBM Watson AI Lab  \\ \textsuperscript{\rm 3} Stanford University \\ \textsuperscript{\rm 4} Massachusetts Institute of Technology 
}

\begin{document}
\maketitle

\begin{abstract}
Humans can abstract prior knowledge from very little data and use it to boost skill learning. In this paper, we propose routine-augmented policy learning (RAPL), which discovers routines composed of primitive actions from a single demonstration and uses discovered routines to augment policy learning. To discover routines from the demonstration, we first abstract routine candidates by identifying grammar over the demonstrated action trajectory. Then, the best routines measured by length and frequency are selected to form a routine library. We propose to learn policy simultaneously at primitive-level and routine-level with discovered routines, leveraging the temporal structure of routines. Our approach enables imitating expert behavior at multiple temporal scales for imitation learning and promotes reinforcement learning exploration. Extensive experiments on Atari games demonstrate that RAPL improves the state-of-the-art imitation learning method SQIL and reinforcement learning method A2C. Further, we show that discovered routines can generalize to unseen levels and difficulties on the CoinRun benchmark.
\end{abstract}
\section{Introduction}
Extensive evidence from cognitive psychology and neuroscience suggests that humans are remarkably capable of abstracting knowledge from very few observations to boost practice in new scenarios. For instance, behavioral experiments on the Atari games \citep{humanLearningInAtari} have demonstrated that human game players could learn from a video of one episode and earn more than double scores than those who do not watch the video. On the contrary, previous Learning from Demonstrations (LfD) approaches either require a large amount of pre-collected data \citep{BehaviorCloning}, an active oracle \citep{DAgger}, or a family of similar tasks \citep{CompILE}. In this paper, we focus on the following question: how can a single demonstration promote policy learning?

Two challenges exist when learning from a single demonstration. First, the agent would often drift away from the few seen expert observations and not return to demonstrated states. Second, high-dimension value function approximators such as deep neural networks \citep{dqn} may over-fit the few demonstrated state-action pairs and cannot overcome unseen environment dynamics. We propose to abstract routines from the demonstration via a non-parametric algorithm and use the routines to help policy learning to address these problems. This idea can alleviate the out-of-distribution problem because routines force the agent to follow segments of the demonstration. Besides, the process of decomposing the demonstration is non-parametric, making the learned policy generalizable to unseen states.

\begin{figure}[t]
\centering
\includegraphics[width=1.0\columnwidth]{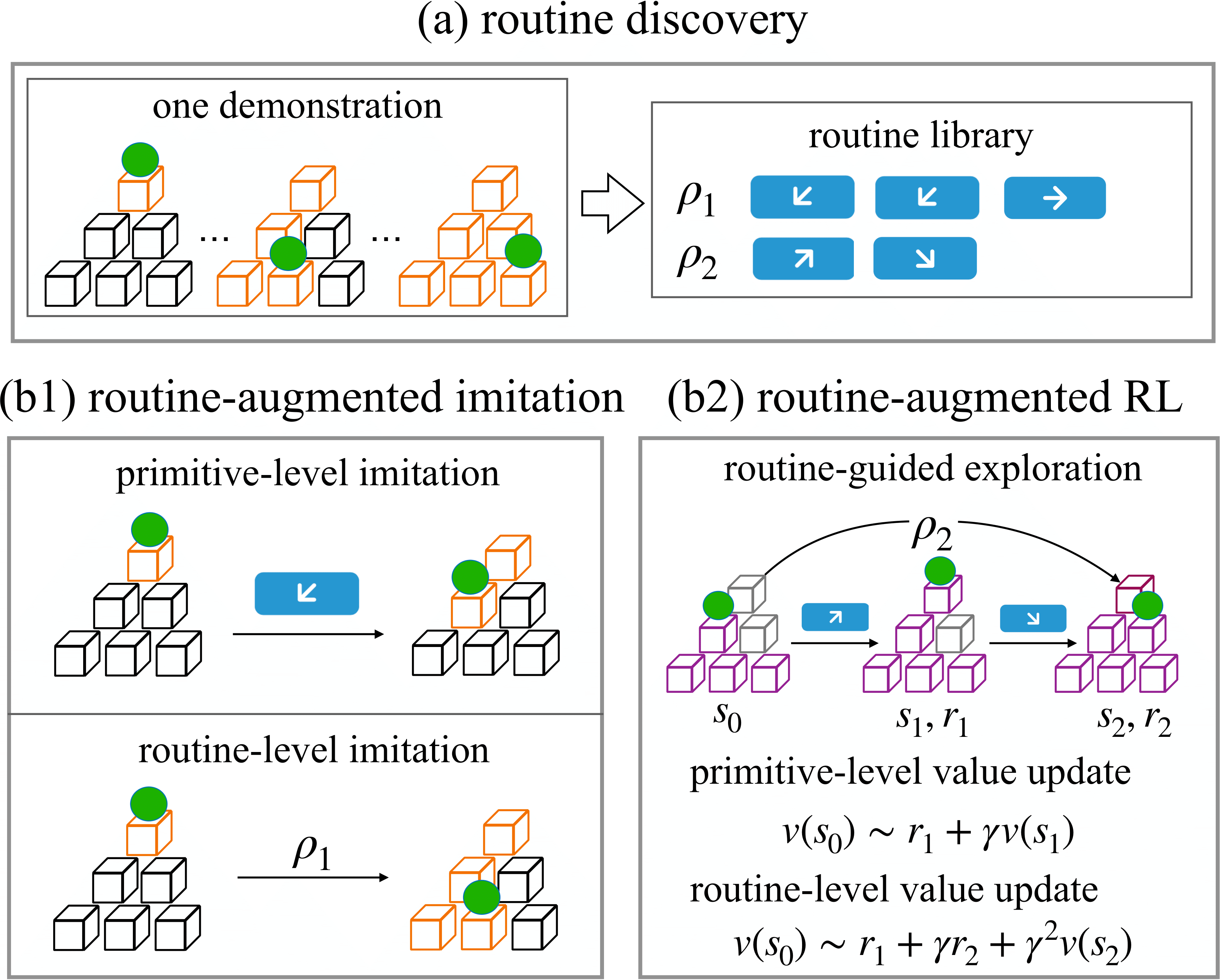}
\caption{Schematic of routine-augmented policy learning (RAPL). In the examples, the green ball represents an agent, which needs to step on every square to change its color (a mini version of Qbert from \citet{atari}). a: We propose to discover a library of routines from a single demonstration. The abstracted routines can be applied to augment both imitation learning and reinforcement learning. b1: For imitation learning (no reward signal), the discovered routines can help the agent imitate the expert's behavior at multiple temporal scales. b2: For reinforcement learning (with reward signal), routines can help exploration as policy shortcuts. The experiences from routine execution are fully exploited to conduct value approximation at both the routine level and the primitive level.}
\label{Fig:teaser}
\end{figure}

The overview of the proposed approach is shown in Figure \ref{Fig:teaser}. A library of routines that represent useful skills is abstracted from the demonstration. The routines can be used in two settings. First, the agent could imitate expert behaviors at multiple temporal scales without access to the reward signal. Second, in reinforcement learning, the abstracted routines can promote deeper exploration and long-range value learning. However, previous option learning approaches must rely on reward signals \cite{OptionCritic, OptionsLearning, SuttonSMDP}. 

We propose a two-phase model for routine discovery. During the first phase, we adopt a non-parametric algorithm, Sequitur \citep{Sequitur}, to discover the structure of the demonstration. Each element in the structure is treated as one routine proposal. In the second phase, we select the best proposals by the frequency and lengths of routine candidates to form a routine library. Too similar routine candidates are pruned to keep the parsimony of the routine library. This model can effectively discover routines without a time-consuming training procedure.

The discovered routines are then used as higher-level actions to boost exploration and policy learning. A na\"ive approach is to run an off-the-shelf policy learning algorithm based on the augmented action space composed by routines and action primitives \citep{DeepRLWithMacroActions, ConstructionMacroActions}. The problem of such an approach is that it ignores the inner structure of routines, so experiences from routine execution are exclusively used to update values at the routine level, which would slow down value learning at the primitive level. Such conflict turns to be a bigger issue as the number of routines grows. To address this problem, since the routines are temporally decomposable, we reuse routine execution experiences to update the value function at the primitive level. Our approach harmonizes the relationship between routines and primitives and has stronger performance when utilizing more and longer routines.

This paper's main contribution is \textit{routine-augmented policy learning} (RAPL): an approach to discover routines from a single demonstration and use them to augment policy learning. Through extensive experiments on the Atari benchmark \citep{atari}, we find that our approach can improve both A2C \citep{AsynchroDeepRL} and SQIL \citep{SQIL} on most of the games. Moreover, we conduct generalization experiments on CoinRun \citep{Coinrun} and observe that the abstracted routines can successfully generalize to unseen levels and harder cases. Our code is now available at https://github.com/sjtuytc/AAAI21-RoutineAugmentedPolicyLearning.

\section{Related Work}

\myparagraph{Imitation Learning} The goal of imitation learning is to learn a policy from the demonstration \citep{LfDsurvey}. Behavior Cloning \citep{BehaviorCloning} only succeeds with a large amount of data. To efficiently leverage the demonstrations, GAIL \citep{GAIL} utilizes adversarial training to prioritizes the demonstration over others. Our approach is different from those approaches because they do not consider discovering higher-level actions from the demonstrations. Besides, we only assume access to one demonstration and need neither a large number of demonstrations nor a family of similar tasks \citep{OneShotImitationLearning}.

\myparagraph{Demonstrations Guided RL} Reinforcement Learning (RL) requires huge time costs and extensive sampling to learn a good strategy \citep{EfficientExploration, CuriosityExplorationSelfSupervised}. Since humans may have prior knowledge of the given task \citep{BayesianPolicySearch}, much recent work \citep{DqfD, DDPGfD, POfD, nair2018overcoming} proposes to leverage demonstrations to help RL. These methods add extra costs in policy learning to penalize the deviation between the learned and the expert policy. Another approach \citep{LearningMontezumaFromOne} utilizes one demonstration to play Montezuma's Revenge, a hard exploration game, by resetting the agent to states in the demonstration. These methods have not considered discovering routines from the demonstration. Moreover, DQfD-based \citep{DqfD, DDPGfD, POfD} approaches assume access to reward signals, while our proposed algorithm can also improve imitation learning from one demonstration.

\myparagraph{Macro-Actions} Macro-actions are temporally extended actions built on primitive actions. In robotics, the classical STRIPS system \citep{STRIPS, MORIS, REFLECT, LearningPlan1972, MacroEmpiricalAnalysis1998} uses predefined \textit{routines} to accelerate making plans. Notably, a few concurrent works consider the discovery of macro actions from the agent's good experiences \citep{ConstructionMacroActions, GrammarActions, CompressionMacro}. Our work is different from them in several folds. First, they adopt an off-the-shelf RL algorithm to train over an action space of macro-actions and primitives. But we propose an efficient and sound manner to train a routine policy. Besides, we propose using routines to augment imitation learning, but they only study adopting macro-actions under reinforcement learning. Third, we do not require the knowledge or the approximation of the environment dynamics, different from \citet{CompressionMacro}. We compare to \citet{DeepRLWithMacroActions} in experiments.

\myparagraph{The Option Frameworks} 
Our work is also related to literature under the \textit{option framework} which learns \textit{options} specified by an initialization set, an intra-option policy, and a termination condition \citep{LearnMacroActions1999, HRLReview2003, OptionCritic, LaplacianOption2017, LearningAbstractOptions, HierarchicalDeepRL, hierarchicalImitationRL}. Our idea of learning at multiple temporal scales originates from Hierarchical Reinforcement Learning \citep{HierarchicalDeepRL}, which jointly learns a meta-controller over options and bottom-level modules to achieve the targets specified in each option. No demonstrations are involved in this work. PolicyBlocks \citep{PolicyBlocks} attempts to discover reusable options from optimal policies. However, it requires a family of tasks to discover options. Some recent work \citep{MultiDDO, DDCO, CompILE, DiscoverMotorPrograms} proposes to discover options from demonstrations and train a controller upon abstracted options. Unlike options adopted in these approaches, our routines are state-independent, and we leave the job of connecting the state with higher-level actions to the phase of policy learning. Furthermore, learning sub-task policies would consume a large number of demonstrations to overcome unseen dynamics, while our approach requires only a single demonstration. We compare to two option learning baselines (ComPILE \citep{CompILE} and OptionCritic \citep{OptionCritic}) in our experiments.
\section{Routine-Augmented Policy Learning (RAPL)}

\subsection{Model Basic}
\newcommand{\MDP}{\Gamma}
\newcommand{\MDPS}{\mathcal{S}}
\newcommand{\MDPA}{\mathcal{A}}
\newcommand{\MDPT}{\mathcal{T}}
\newcommand{\MDPR}{\mathcal{R}}
\newcommand{\MDPDIS}{\gamma}
\newcommand{\MDPPOLICY}{\pi}
\newcommand{\MRDPL}{\mathcal{L}}
\newcommand{\MixedAction}{ \widetilde{\rho}}
\newcommand{\LengthRt}{ {\left|\rho \right|}}
\newcommand{\LengthMixed}{ {\left|\widetilde{\rho}\right|}}
\newcommand{\AugmentedActionSpace}{\widetilde{\mathcal{L}}}

\myparagraph{MDPs} During a timestep $t$ on an Markov Decision Process (MDP) $\MDP$, the agent chooses an action $a_t$ from a predefined primitive action set $\MDPA$ after receiving an observation state $s_t \in \MDPS$. The environment provides a transition function $\MDPT(s_t, a_t)$, a reward $r_t$ (not available in imitation learning), and a discount factor $\MDPDIS$. The core problem of an MDP is to find a policy function $\MDPPOLICY(a_t|s_t)$. In this paper, we focus on MDPs with high-dimensional states and discrete actions.

\myparagraph{Routines and Routine Policies} We define a routine $\rho$ to be a sequence of primitive actions $(a^{(1)}, a^{(2)}, ..., a^{(\LengthRt)})$ and $\LengthRt$ to be its length. The notion of \textit\text{routine} appeared in \citet{STRIPS} and we emphasis that routines are abstracted from demonstrations in this paper (different from hand-crafted macro actions). A \textit{routine library} $\MRDPL$ is defined to be a set of discovered routines for a task. After routines are introduced, an agent can choose one routine $\rho_t \in \MRDPL$ or a primitive action $a_t\in \MDPA$ based on a state $s_t \in \MDPS$. When a routine $\rho_t$ is chosen, the primitive actions in $\rho_t$ are executed sequentially, and the agent would make the next decision after the execution of $a^{(|\rho_t|)}$. For convenience, we use $\AugmentedActionSpace=A\cup \MRDPL$ to represent the routine-augmented action space and $\MixedAction \in \AugmentedActionSpace$ to represent an \textit{extended routine}. Plus, we define $\left|\MixedAction \right|$ to be the length of $\MixedAction$ (the length of a primitive action is one). The goal is to find a \textit{routine policy} $\pi(\MixedAction_t|s_t)$, which specifies the distribution of extended routines for a state at timestep $t$.

\subsection{Routine Discovery}
\label{subsec:RoutineAbstraction}
We propose a two-phase algorithm for routine discovery from a single demonstration. During the first phase, we construct a set of routine proposals from the demonstration. After that, we select the best routines from the routine candidates measured by frequency and length. Those selected best routines form a routine library to augment policy learning. The pseudo-code of routine discovery is provided in the supplementary material.

\myparagraph{Routine Proposal}
The key idea is that one can decompose the demonstration and consider each segment as a routine proposal. We adopt a non-parametric algorithm, Sequitur \citep{Sequitur}, to recover the structure of the demonstration. Sequitur takes the demonstrated action trajectory as input and outputs a context-free grammar generating the whole action sequence. The grammar is represented as a set of rules. Each rule in the grammar connects from a variable to a sequence of variables. Sequitur introduces intermediate variables, each of which can be transferred to a sequence of terminal variables (variables that do not connect to any variables in the rules). Each terminal variable corresponds to a primitive action in the demonstrated action sequence. Therefore, each intermediate variable can be considered as a routine candidate. We refer readers to \citet{Sequitur} for more details about Sequitur.

\myparagraph{Routine Selection} 
After acquiring the routine candidates, we use a selection procedure to limit the routine library's size to be $ K $, a hyper-parameter. We adopt a hybrid metric, considering both the frequency and length of the routine proposals. On the one hand, routines frequently appear in the demonstration may entail useful skills to conquer tasks. On the other hand, we encourage selecting longer routines to encode more expert policy patterns. Denote the occurrence time of one routine $\rho$ in the demonstrated action sequence to be $f(\rho)$, and its length to be $\LengthRt$. The score of a routine can be written as $f(\rho) + \lambda^\text{length} \LengthRt$, where $\lambda^\text{length}$ is a balancing factor. To prevent introducing too many similar routines, we only leave the routine with the highest score when similar routines are detected. The similarity is measured by the Levenshtein distance \citep{LevenshteinDistance}, which is the edit distance of two sequences. Finally, the $K$ routine candidates with the highest scores are selected to form a routine library.

\subsection{Routine Policy Learning}
\label{subsec:routine_policy_update}
After introducing routine library $\MRDPL$, the agent's action space becomes $\AugmentedActionSpace=A\cup \MRDPL$. One na\"ive approach is to regard routines as black-box actions and use an off-the-shelf policy learning algorithm to train an agent with the augmented action space $\AugmentedActionSpace$ \citep{DeepRLWithMacroActions, CompressionMacro}. Such an approach fails to consider the temporal structure of routines and would slow down policy learning when $\AugmentedActionSpace$ consists of more and longer routines. We propose to reuse experiences at multiple temporal scales to update policy efficiently.

We instantiate this idea in two settings. On the one hand, when the reward is not available, routines are used to augment SQIL \citep{SQIL}, a state-of-the-art imitation learning algorithm, to enable imitation learning over multiple temporal scales. On the other hand, we use routines to promote the standard reinforcement learning algorithm A2C. We formulate the learning targets for those two algorithms in the following paragraphs.

\myparagraph{RAPL-SQIL} SQIL \citep{SQIL} is a recently proposed simple yet effective imitation learning approach. It gives all the experiences from the demonstration a constant reward $r=1$. Besides, all the newly explored experiences are given a reward $r=0$. This can encourage the agent to go back to the demonstrated states. The demonstration is represented as $D_\text{prim}$, where each element in $D_\text{prim}$ is a tuple $(s_t, a, s_{t+1})$. We find all the occurrences of every discovered routine $\rho \in \MRDPL$ in the demonstrated action sequence. Combining each occurrence with the states before and after routine execution in the demonstration, we get a higher-level demonstration $D_\text{routine}$. Each entry in $D_\text{routine}$ is represented as $(s_t, \rho, s_{t + \LengthRt})$, where $s_t$ and $s_{t + \LengthRt}$ are the states before and after the execution of $\rho$ correspondingly. Therefore, $D_\text{routine}$ and $D_\text{prim}$ 
contain experiences in routine-level and primitive-level accordingly. The squared soft Bellman error is given as
\begin{equation}
\begin{aligned}
\delta^{2}(\mathcal{D}, r) & = \frac{1}{|\mathcal{D}|} \\ &  \sum_{\left(s_t, \MixedAction, s_{t + \LengthMixed}\right) \in \mathcal{D}} 
\left(Q_{\theta}(s_t, \MixedAction)- Q_\text{target}( \MixedAction, s_{t + \LengthMixed}, r)\right)^{2},
\end{aligned}
\end{equation}
\begin{equation}
\begin{aligned}
Q_\text{target}(\MixedAction, s_{t + \LengthMixed}, r)  & = R_{sq}(\MixedAction, r) + \\ & \Gamma(\MixedAction)\log  \left( \sum_{\MixedAction' \in \AugmentedActionSpace}\exp \left(Q_{\theta}\left(s_{t + \LengthMixed}, \MixedAction'\right)\right)\right),
\end{aligned}
\end{equation}
where $R_{sq}(\MixedAction, r)$ and $\Gamma(\MixedAction)$ are the reward function and the discount factor defined for the extended routine $\MixedAction$. Since the execution of routines connects two states with an interval of $\left| \MixedAction \right|$, we define the extended routine's reward function to be the sum of discounted primitive rewards and its discount factor to be $\lambda$ discounted by $\left| \MixedAction \right|$ times. Formally,
\begin{equation}
R_{sq}(\MixedAction, r) = \sum_{\tau=1}^{\left|\MixedAction\right|}\gamma^{\tau - 1}r, \qquad\Gamma(\MixedAction) = \gamma ^ {\left| \MixedAction \right|}.
\end{equation}
The final loss of SQIL with routines is 
\begin{equation}
\mathcal{L}^{\mathcal{SR}} = \delta^{2}(\mathcal{D}_\text{prim}\cup \mathcal{D}_\text{routine} , 1) + \lambda_\text{sample} \delta^{2}(\mathcal{D}_\text{sample}, 0),
\end{equation}
where $D_\text{sample}$ represents the collected experiences of interactions with the environments and $\lambda_\text{sample}$ is the balancing hyperparameter between the demonstrated and explored transitions.

\myparagraph{RAPL-A2C} 
We apply the augmented action space to a state-of-the-art reinforcement learning method Advantage Actor Critic (A2C)~\citep{AsynchroDeepRL}. A2C with routines learns a policy function $\pi(\MixedAction_t|s_t;\theta_\pi)$ and a state value function $V(s_t; \theta_v)$. We compute two advantage functions to backtrack delayed rewards to the current state, differing in their temporal granularity. In the first advantage function $A_\text{routine}$, we compute the return from $N$-step of routine experiences. Denote the explored on-policy experiences of routine execution to be $\{(s_{t_\tau}, \MixedAction_{t_\tau}, R_{t_\tau}, s_{t_{\tau + 1}}) | 0 \leq \tau \leq N -1\}$, where $t_i=t_0 + \sum_{\tau=0}^{i-1}|\MixedAction_{t_\tau}|$. Note the total primitive steps are $\sum_{\tau=0}^{N-1}|\MixedAction_{t_\tau}|$, which could be much larger than $N$. The reward of a routine is the sum of discounted primitive rewards, so we have $R_{t_i}=\sum_{\tau=t_{i}}^{t_{i+1} - 1}\gamma^{\tau - t_i} r_\tau$. Then we can write the routine-level advantage function as
\begin{equation}
\label{eq:routine_level_ad}
A_\text{routine}=\sum^{N-1}_{i=0}\gamma^{t_i - t_0}R_{t_i} + \gamma^{t_N-t_0} V(s_{t_N}) - V(s_{t_0}).
\end{equation}

In the second advantage function, we take care of the primitive-level value approximation and compute $N$-step bootstrapping for primitives. From the experiences of routine execution, we randomly sample an $N$-step consecutive primitive experience, represented as $\{(s_{\tau}, a_{\tau}, r_{\tau}, s_{\tau+1}) |t_j \leq \tau \leq t_j + N - 1\}$ (note that we can get access to the intermediate states during routine execution). Then we give the primitive-level advantage function as 
\begin{equation}
\label{eq:prim_level_ad}
A_\text{prim}=\sum^{N-1}_{i=0}\gamma^{i}r_{t_j+i} + \gamma^{N} V(s_{t_j+N}) - V(s_{t_j}).
\end{equation}

To optimize the policy function, we pose a policy gradient loss and an entropy loss:
\begin{align}
\mathcal{L}^\text{policy}&=-A_\text{routine} \log\pi(\MixedAction_{t} | s_{t_0}; \theta_\pi),\\
\mathcal{L}^\text{entropy} &= \sum_{\MixedAction}\pi(\MixedAction | s_{t_0}; \theta_\pi)\log\pi(\MixedAction | s_{t_0}; \theta_\pi).
\end{align}
The final loss for A2C with routines is
\begin{equation}
\begin{aligned}
\mathcal{L}^{\mathcal{AR}} & = \mathbb{E}(\mathcal{L}^\text{policy}  + \lambda^\text{entropy}\mathcal{L}^\text{entropy} \\ & + \lambda^\text{value}(\left\|A_\text{routine}\right\|^2 + \lambda^\text{prim}\left\|A_\text{prim}\right\|^2)),
\end{aligned}
\end{equation}
where the expectation is taken over all sampled experiences. We denote $\lambda^\text{value}$, $\lambda^\text{prim}$, $\lambda^\text{entropy}$ to be the balancing factors for each loss term.

\begin{figure}[t]
\centering
\includegraphics[width=1.0\columnwidth]{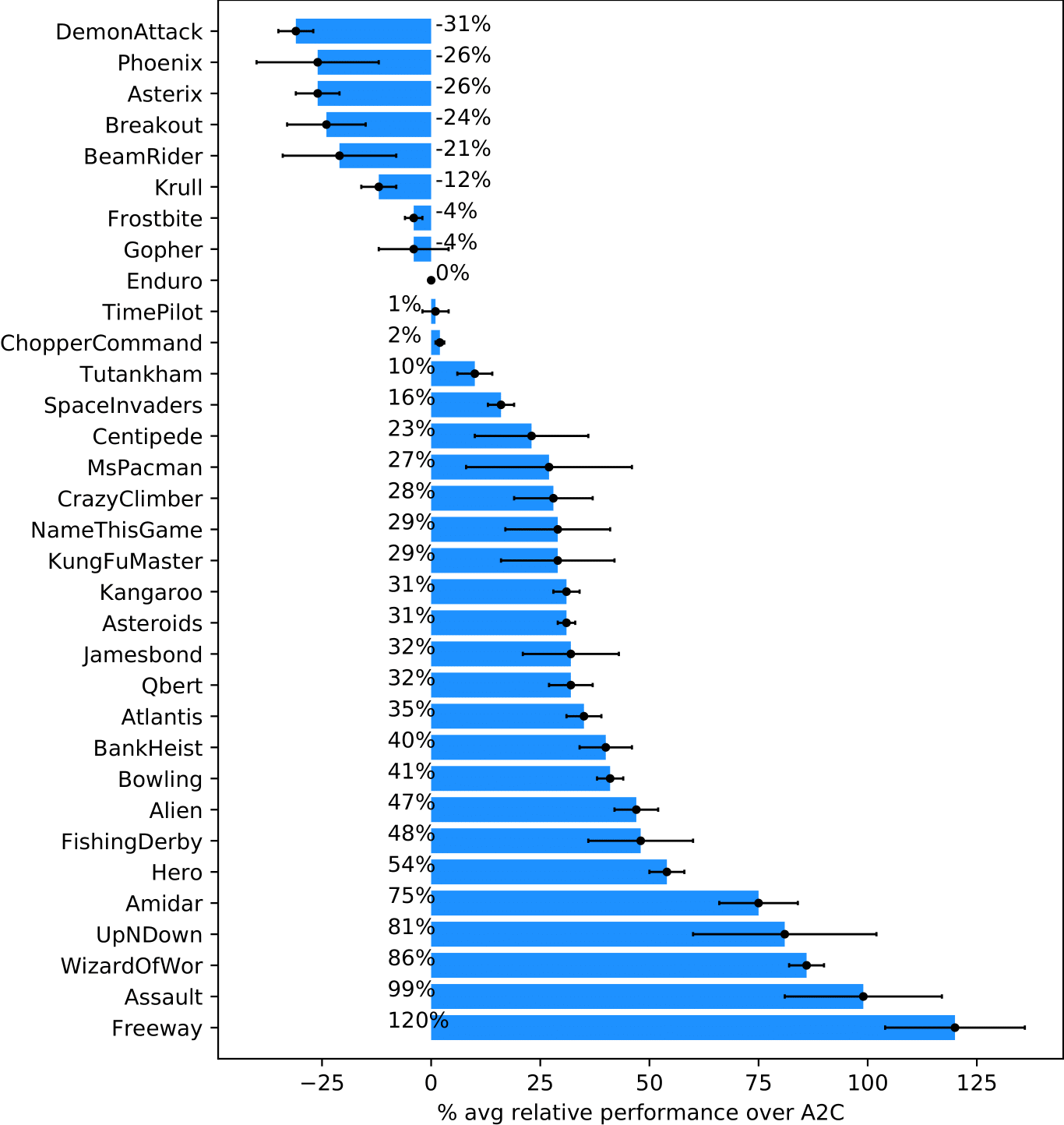}
\caption{Relative performance of RAPL-A2C over A2C on Atari. Denote $S_{R}$ as the score of RAPL-A2C and $S_A$ is the score of A2C. The relative performance is calculated by $(S_{R} - S_A) / |S_A| \times 100\%$. Each number is averaged over five random agents and we also plot the stand error of the numbers.}
\label{Fig:results_atari}
\end{figure}

\begin{figure*}[t]
\centering
\includegraphics[width=1.0\textwidth]{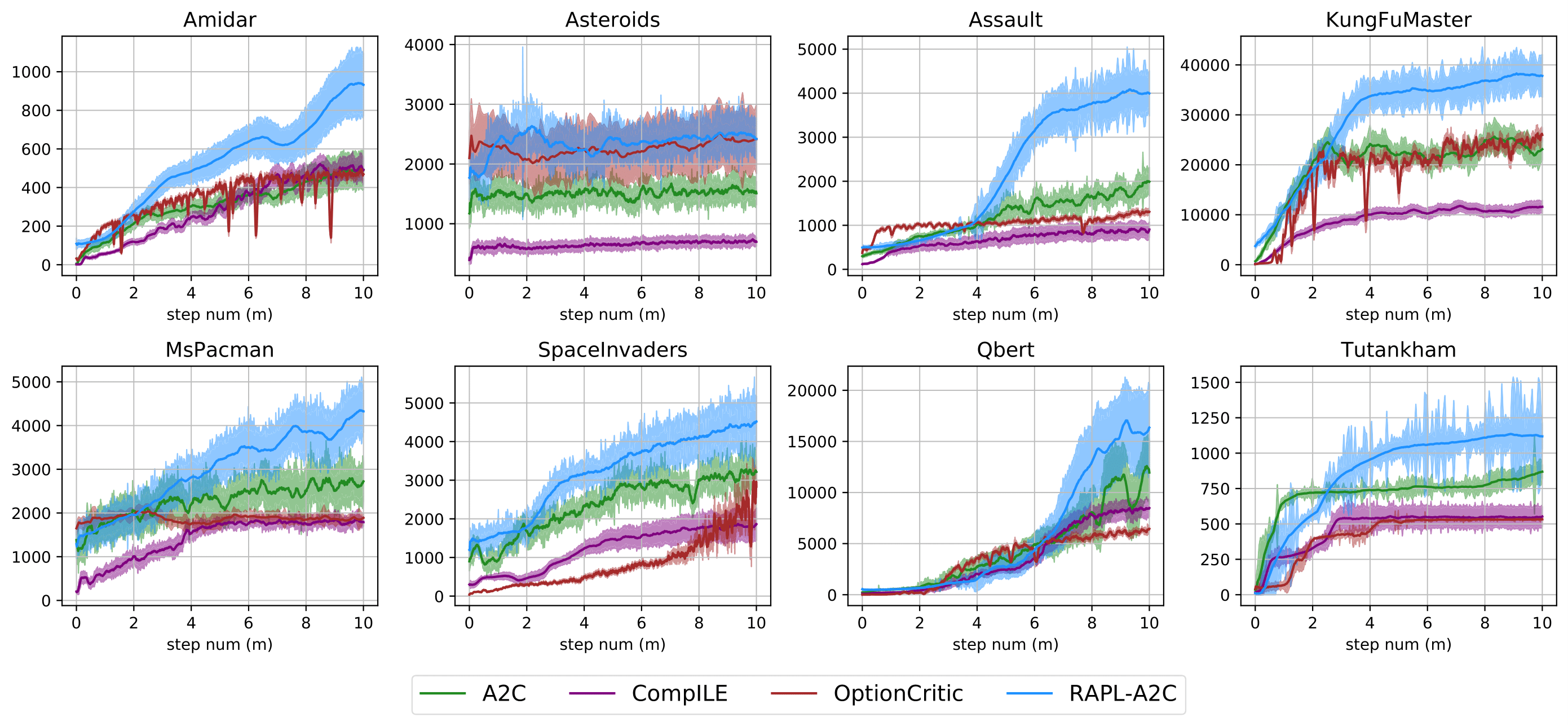}
\caption{Training curves on eight randomly selected Atari games in comparison with several RL baselines. We plot both the mean and standard deviation in those curves across five agents with random seeds.}
\label{Fig:CurveAtari}
\end{figure*}

\begin{figure}[t]
\centering
\includegraphics[width=1.0\columnwidth]{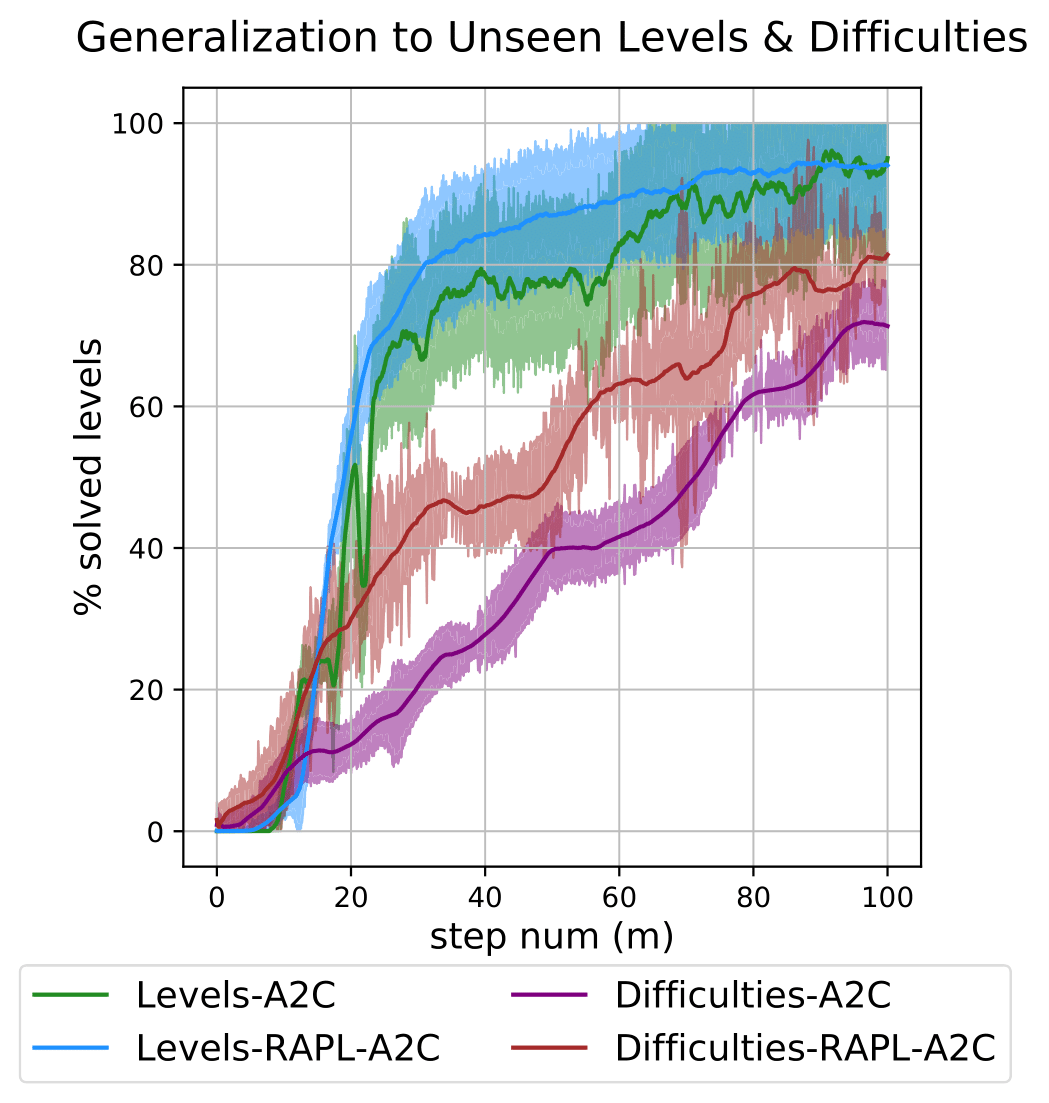}
\caption{Generalization curves on CoinRun. We use "Levels" and "Difficulties" to indicate generalization to unseen levels and difficulties accordingly. We show both the mean and the standard deviation across five random seeds.}
\label{Fig:CurveCoinrun}
\end{figure}

\begin{figure}[t]
\centering
\includegraphics[width=1.0\columnwidth]{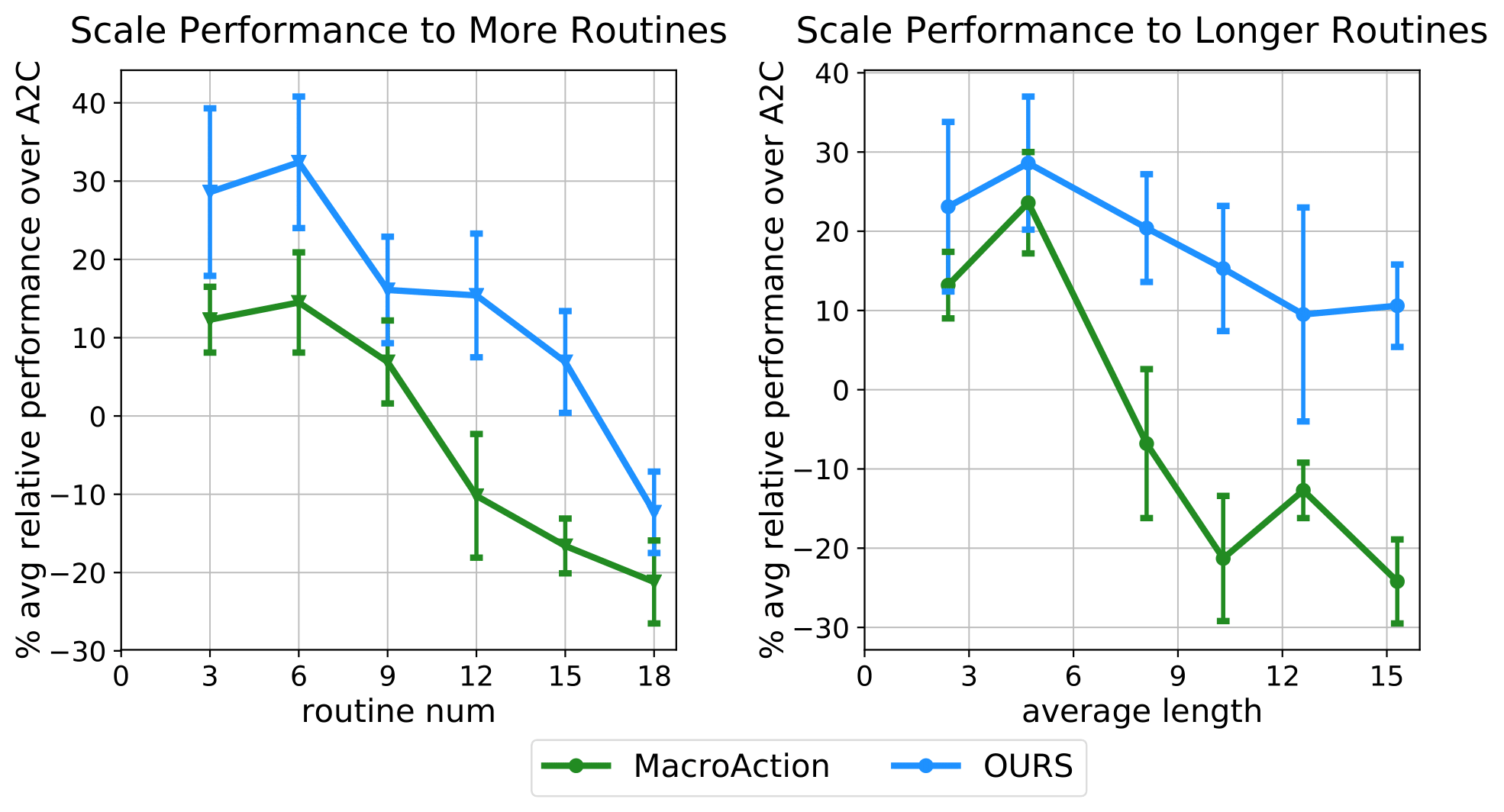}
\caption{The scalability of our approach on Atari games. Each number represents the relative performance over A2C averaged on 33 Atari games. Mean and standard error over five random agents are shown in the figure.}
\label{Fig:scale_performance}
\end{figure}

\begin{figure}[t]
\centering
\includegraphics[width=1.0\columnwidth]{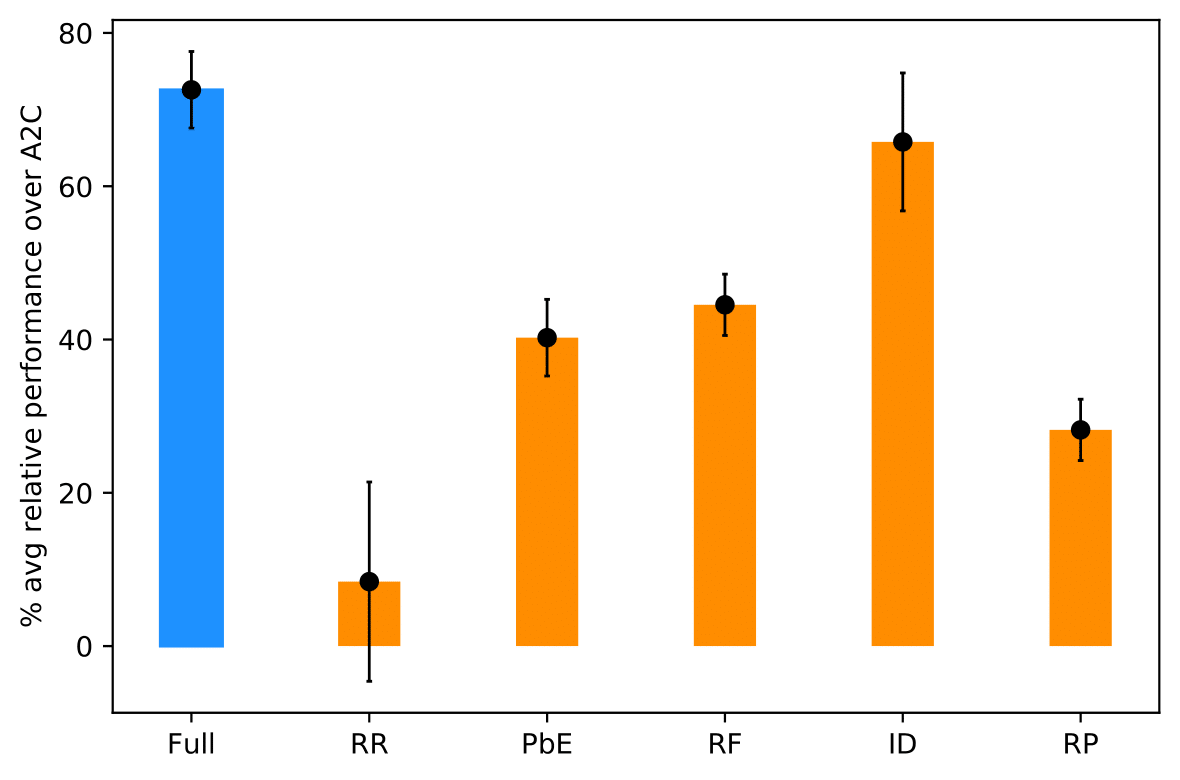}
\caption{Comparison of ablated routine discovery models on Atari games. Mean and standard error over five random agents are shown in the figure.}
\label{Fig:ablation_routine_discovery}
\end{figure}

\begin{table}[t]
\centering
\begin{tabular}{l|c|c}
\toprule
              & Alignment ($\pm$ std)               & Mean ($\pm$ std)                    \\ \midrule
BC            & 0.18 ($\pm$ 0.03)                   & 18.3\% (2.1\%)                        \\
GAIL          & 0.16 ($\pm$ 0.08)                   & 26.4\% (1.6\%)                        \\
SQIL          & 0.28 ($\pm$ 0.07)                   & 29.4\% (3.2\%)                        \\ \midrule
\textbf{RAPL-SQIL} & \textbf{\textbf{0.34} ($\pm$ 0.07)} & \textbf{36.1\% ($\pm$ 3.6\%)} \\ \bottomrule
\end{tabular}
\caption{Comparing with several imitation learning baselines on 33 Atari games. We shown both alignment scores (defined in Eq.~\ref{eq:alignment_score}) and mean of human-normalized scores \cite{dqn} which indicates the alignment performance with regarding to the demonstration. Each number in the table is averaged over five random seeds.}
\label{tab:imitation_learning}
\end{table}

\section{Experiments}
We investigate the following questions by experiments: 1) Does RAPL improve imitation learning and reinforcement learning methods? 2) Does our approach outperform other baselines to learn from demonstrations? 4) How does our approach perform when scaling to more and longer routines? 4) Can discovered routines generalize to unseen scenarios?

\subsection{Experimental Setting}
\myparagraph{Environment Description} Our experiments are conducted on the Atari benchmark \citep{atari} and CoinRun \citep{Coinrun}. We use $33$ Atari games selected by \citet{FiGAR} (including all the games in their experiments expect for Koolaid that is not supported in our experiment platform Gym \citep{Gym}). We use a frame-skip of $4$, a frame-stack of $4$, and the minimal action space \citep{atari}. We use the convolutional neural network described in \citet{dqn} on Atari games. CoinRun is a recent benchmark that has different levels that enable quantifying the generalization ability of RL methods. It also provides two difficulties modes: easy and hard. We adopt a minimal action space composed of \texttt{Left, Down, Up, Right, Nope} for the convenience of presentation. We do not paint velocity information in the observation. No frame-stack is used in CoinRun as in \citet{Coinrun}. For CoinRun, we use the IMPALA-CNN architecture \citep{IMPALA}. All the environmental settings are kept the same for all approaches to ensure fairness.

\myparagraph{Demonstration Collection} For all the games, we only use one demonstration generated by a trained A2C agent. We use $\lambda^\text{value}=0.5$ and $\lambda^\text{entropy} = 0.01$ to balance the value loss and entropy loss accordingly. We set $\lambda^\text{prim} = 1.0$ when using routine augmentation. The optimizer is RMSProp with a learning rate $7 \times 10^{-4}$, a linear decay of $10^{-5}$ per timestep. We use entropy regularization with $\beta = 0.02$. The return is calculated for $N=5$ steps. Each agent is trained for $10$ million steps. 

\myparagraph{Routine Discovery} In all experiments, we set the balancing factor between frequency and length to be $\lambda^\text{length}=0.1$. Moreover, the number of routines is set to $K=3$. We would leave the best routine between routines whose Levenshtein distance is smaller than $\alpha=2$. These hyper-parameters are coarsely selected by validating on a few games (refer to Supplementary for details), and they are kept all the same for all the other games.


\subsection{Imitation Learning with Routines}

We validate whether discovered routines can improve SQIL \citep{SQIL} and compare our results with Behavior Cloning (BC) \citep{BehaviorCloning}, which conducts supervised learning from demonstration data without any environment interaction. Moreover, we compare with a standard model-free imitation learning algorithm GAIL \citep{GAIL}. We thank the author of SQIL \citep{SQIL} for providing the implementation of these algorithms. As described in \citet{SQIL}, we use $\lambda^\text{sample} = 1$. The optimizer is Adam \citep{adam} with a learning rate $10^{-3}$. The agent is trained via $10^5$ on-policy rollouts. Each score reported is the average reward on $100$ episodes after training.

We propose a metric of alignment score to measure how well the imitator imitates the expert. Given the demonstrated action trajectory $\iota_{d}$ and the action trajectory produced by the trained agent $\iota_{t}$ (note $\iota_{t}$ is padded or cut to have the same length with $\iota_{d}$), we compute the alignment score $s$ as 
\begin{equation}
\label{eq:alignment_score}
s = 1 - \frac{D(\iota_{d}, \iota_{t})}{|\iota_{d}|},
\end{equation}
where $D$ is the Levenshtein distance and $|\iota_{d}|$ denotes the length of the demonstration. 

We present the results in Table~\ref{tab:imitation_learning}. We notice that RAPL-SQIL could help the agent perform in line with the demonstration. The agent effectively learns when to use routine through a single demonstration and environmental interactions. The results indicate that routines can effectively force the agent to follow the patterns of the single demonstration. Besides, this fact suggests that imitating expert's policy at multiple temporal scales would enhance imitation learning.

\subsection{Reinforcement Learning with Routines} We first study whether routine discovery can improve model-free reinforcement learning method A2C \citep{AsynchroDeepRL}. We then compare with a recent proposed parametric routine discovery approach ComPILE \citep{CompILE}. ComPILE first decomposes the demonstration into segments via a parametric recognition model; it then trains sub-policy and the termination condition for each segment via supervised learning. After that, it trains an A2C controller over an augmented space composed of those segments and primitives. We further compare to an option learning baseline, OptionCritic \citep{OptionCritic}, which is also based on the actor-critic architecture and uses the two-layer optimization of Bellman targets. For all the agents trained with A2C, we use the same hyper-parameters used in expert training.

We list the relative performance of routine-augmented A2C over A2C in Figure~\ref{Fig:results_atari}, which indicates that our approach achieves the same or better performance in $25$ out of $33$ games. This fact indicates that the routines discovered from the demonstration can effectively enhance the exploration of reinforcement learning. The training curves of comparison on Atari games are shown in Figure~\ref{Fig:CurveAtari}. Our approach outperforms baselines on most of the games. We notice that ComPILE usually deteriorates the baseline of A2C. The first reason for this is that ComPILE requires many demonstrations from a family of tasks to train the sub-task policies and termination conditions. When only a single demonstration of a task is given, those parametric policies and conditions cannot generalize to unseen states.  The OptionCritic does not use the demonstration but uses more parameters to model the option policy, intra-option policy, and termination conditions. Therefore, it achieves relatively limited performance gain over A2C. In contrast, our proposed approach successfully discovers effective routines from a single demonstration, which further generalizes to states that are not seen from the demonstration.

\subsection{Scalability of RAPL}
We study the performance of RAPL when scale to more or longer routines in comparison with a na\"ive baseline MacroAction \citep{DeepRLWithMacroActions}. MacroAction appends routines into the agent's action space and adopts an off-the-shelf A2C algorithm to train the controller. To ensure fairness, we adopt the same routines discovered from the demonstration for MacroAction.

The results are shown in Figure~\ref{Fig:scale_performance}. Our approach performs better on more and longer routines. The first reason is that MacroAction does not reuse the experience from routine execution to update the value function at the primitive-level as in Eq.~\ref{eq:prim_level_ad}. So when using longer routines, the value function's bootstrap involves too many primitive steps (they do not interrupt during the execution of routines \citep{SuttonSMDP}). Therefore the value estimation of middle states during execution is less accurate, leading to inferior performance. When using more routines, RAPL-A2C can efficiently share experiences of routines to primitives, so more routines deteriorate the performance to a less extent. Furthermore, it does not take care of the temporal discount relationship when the execution of routines triggers temporal abstraction. For example, it defines the reward of a routine execution to be the sum of rewards during its execution, which contradicts to Eq.~\ref{eq:routine_level_ad}.

\subsection{Effectiveness of Routine Discovery}

We compare the full model (Full) to the following ablated versions to validate routine discovery effectiveness. Each model is tested on eight Atari games listed in Figure~\ref{Fig:CurveAtari}. (1) Random Routines (RR), where each routine is generated randomly. (2) The proposal by Enumeration (PbE) where we enumerate all the possible combinations of primitive actions to form routine candidates. (3) Random Fetch (RF) where we random fetch sub-sequences from the demonstration to form routines. (4) Imperfect Demonstration (ID) where the expert is only trained with 1 million steps. (5) Repeat (RP), where the routines are the repetition of most frequently used atomic actions in the demonstration \citep{FiGAR}.

Despite the specified ablated component, other details are the same as the full model (including the number and the length of each routine). We run each model for five random seeds and report both the mean and standard deviation in Figure~\ref{Fig:ablation_routine_discovery}. We observe that ablating any of the components would harm the performance of discovered routines. Random Routines and Proposal by Enumeration perform worst among all the models because they do not leverage the demonstration's information and only select routines by the heuristic. The inferior performance of Random Fetch suggests it is beneficial to propose routines via Sequitur. Our model also outperforms simply repetition. We can also find that our approach is robust to imperfect demonstrations because useful skills exist in the imperfect experts.

\subsection{Generalization of Routines}
We conduct various experiments on CoinRun to validate the generalization ability of RAPL. We train two agents by both A2C and RAPL-A2C on the same $100$ easy levels. Then we test them on $100$ unseen easy levels to test the generalization ability to unseen levels. After that, we test both agents on 100 hard levels to test the generalization ability across difficulties. 

The results are shown in Figure~\ref{Fig:CurveCoinrun}. Both A2C and RAPL-A2C fit well in the training levels. Notably, we find RAPL-A2C improves generalization ability. On the one hand, we observe that the discovered routines can successfully generalize to unseen levels. On the other hand, discovering useful skills from relatively simple domains might also promote policy learning in unseen hard domains. These facts indicate that routines may alleviate over-fitting problems of deep neural networks.

\myparagraph{Visualization of Trained Agents} We provide a visualization of two trained agents in the Supplementary. The discovered routines represent the ability to jump far and high, helping the agent to overcome obstacles. Besides, the policy trained by plain A2C is pretty noisy due to the sparse reward in CoinRun (the agent only gets positive rewards at the end of each episode). Routines regularize the policy towards the optimal policy, which contributes to the improvement in generalization. Finally, we observe that adopting routines can benefit the interpretability of policy since routines are higher-level actions that are easier for a human to understand.

\section{Conclusion}
In this paper, we have presented \textit{routine-augmented policy learning} (RAPL) to discover a set of routines from a single demonstration and augment policy learning via the discovered routines. From extensive experiments on Atari, we found that routines can enhance imitation learning by learning at multiple temporal scales, and routines can promote exploration in reinforcement learning. Besides, from experiments on CoinRun, we found that the discovered routines can generalize to unseen levels and harder domains. We hope that our proposed approach can inspire further work to extend RAPL to continuous action domains. Moreover, discovering routines with rich semantic information would be a promising future direction.
\newpage
\section{Acknowledgements}
This work was supported in part by the Center for Brains, Minds and Machines (CBMM, NSF STC award CCF-1231216), ONR MURI N00014-16-1-2007, MIT-IBM Watson AI Lab, and MERL.

\bibliography{reference}
\end{document}